\pdfoutput=1

\documentclass[11pt]{article}

\usepackage[]{EMNLP2022}

\usepackage{placeins}
\usepackage{multirow}
\usepackage{graphicx}
\usepackage{pgf}
\usepackage{lmodern}
\usepackage{layouts}
\usepackage{hyperref}
\usepackage{mdframed}
\usepackage{lipsum}
\usepackage{amsmath}

\usepackage{times}
\usepackage{latexsym}

\usepackage[T1]{fontenc}

\usepackage[utf8]{inputenc}

\usepackage{microtype}
\usepackage{inconsolata}

\definecolor{Orange}{RGB}{244, 101, 66}

\usepackage{enumitem}
\usepackage{paralist}

\title{Representation Learning for Resource-Constrained Keyphrase Generation}

\author{
Di Wu$^1$, Wasi Uddin Ahmad$^2$\thanks{~ Work done while at UCLA.}, Sunipa Dev$^1$, Kai-Wei Chang$^1$ \\
University of California, Los Angeles$^1$, AWS AI Labs$^2$ \\ 
\texttt{\{diwu,sunipa,kwchang\}@cs.ucla.edu, wasiahmad@ucla.edu}
}

\begin{document}
\maketitle

\setlength{\abovedisplayskip}{5pt}
\setlength{\belowdisplayskip}{5pt}

\begin{abstract}
State-of-the-art keyphrase generation methods generally depend on large annotated datasets, limiting their performance in domains with limited annotated data. To overcome this challenge, we design a data-oriented approach that first identifies salient information using \textit{retrieval-based} corpus-level statistics, and then learns a task-specific intermediate representation based on a pre-trained language model using large-scale unlabeled documents. We introduce \textit{salient span recovery} and \textit{salient span prediction} as denoising training objectives that condense the intra-article and inter-article knowledge essential for keyphrase generation. Through experiments on multiple keyphrase generation benchmarks, we show the effectiveness of the proposed approach for facilitating low-resource keyphrase generation and zero-shot domain adaptation. Our method especially benefits the generation of absent keyphrases, approaching the performance of models trained with large training sets. 
\end{abstract}
\section{Introduction}

Keyphrases of a document are the phrases that summarize the most important information. In the keyphrase generation task, given a document, a model is required to generate a set of keyphrases, each of which can be classified as a \textit{present keyphrase} if it appears as a contiguous text span in the document or an \textit{absent keyphrase} otherwise. The generated keyphrases can facilitate a wide range of applications, such as document clustering \citep{Hammouda2005}, recommendation systems \citep{10.1145/1367497.1367723, 10.1145/1871437.1871754}, information retrieval tasks \citep{10.1145/312624.312671, kim-etal-2013-applying, Tang2017QALinkET, boudin-etal-2020-keyphrase}, text summarization \citep{10.5555/1039791.1039794}, and text classification \citep{10.3115/1220175.1220243, wilson-etal-2005-recognizing, berend-2011-opinion}. 

Recent years have seen promising results of \emph{neural} keyphrase generation approaches as more large-scale annotated training datasets become available \citep{meng-etal-2017-deep,chan-etal-2019-neural,chen-etal-2020-exclusive,yuan-etal-2020-one,ahmad-etal-2021-select,ye-etal-2021-one2set}. For instance, KP20k \citep{meng-etal-2017-deep}, a popular scientific keyphrase generation dataset, contains over 500,000 documents in its training set. Recent datasets in the news, science, or social media domains are often of a similar scale \citep{gallina-etal-2019-kptimes, cano-bojar-2019-keyphrase, yuan-etal-2020-one}. On the other hand, the poor out-of-distribution generalization ability of keyphrase generation models is often observed \citep{chen-etal-2018-keyphrase}. This brings the challenge of training neural keyphrase generation models in the domains where gathering labeled data is difficult (e.g., due to privacy concerns) or domains that evolve as time goes by (e.g., with the creation of new concepts).

\begin{figure}[!t]
    \centering
    \small
    \resizebox{\linewidth}{!}{%
    \begin{tabular}{| p{0.98\linewidth} |}
    \hline \\ 
    \vspace{-12pt} 
    \textbf{Input:} \textcolor{blue}{localization} and \textcolor{blue}{regularization} behavior of \textcolor{blue}{mixed finite elements} for 2d structural problems with damaging material. \textbf{<sep>} a class of lagrangian \textcolor{blue}{mixed finite elements} is presented for applications to 2d structural problems based on a \textcolor{blue}{damage} constitutive model. attention is on \textcolor{blue}{localization} and \textcolor{blue}{regularization} issues as compared with the correspondent behavior of lagrangian displacement based elements. \\ [2pt]
    \hline \\ \vspace{-12pt}  
    \textbf{Present Keyphrases:} \textcolor{blue}{localization }; \textcolor{blue}{regularization }; \textcolor{blue}{ mixed finite elements }; \textcolor{blue}{damage } \\
    \textbf{Absent Keyphrases:} \textcolor{red}{hybrid formulations} ; \textcolor{red}{plasticity} \\ [2pt]
    \hline
    \end{tabular}}
    \caption{An example keyphrase generation case. The input document contains a title and some body text, separated by a separator token \textbf{<sep>}. }
    \label{example-case}
\end{figure}

In this paper, we focus on improving the keyphrase generation performance in such "low-resource" scenarios where annotated data is limited. Pre-trained language models (PLMs), task-specific pre-training, and domain-specific pre-training have successfully driven low-resource NLP applications \citep{zhang2020pegasus, zhang-etal-2020-multi-stage, gururangan-etal-2020-dont, hedderich-etal-2021-survey, zou-etal-2021-low, yu-etal-2021-adaptsum}. These approaches often rely on objectives such as masked language modeling \citep{devlin-etal-2019-bert} or text infilling \citep{lewis-etal-2020-bart} to provide self-supervised learning signals. Can we find similar self-supervision signals for keyphrase generation to make the downstream supervised fine-tuning more data-efficient? 

To fulfill this goal, language modeling based on random masking or infilling may not be optimal. Intuitively, training to recover from random masking via maximum likelihood estimation (MLE) teaches the model to generate probable and coherent output but does not encourage the model to generate key information. For example, given the instance "\texttt{A(n) \_\_\_ approach is what we need}", based on the context, a general language model may fill in with general words such as "\texttt{creative}" or "\texttt{reliable}". By contrast, a model that is better equipped for keyphrase generation may fill in with more specific and salient information, such as "\texttt{multimodal}" or "\texttt{object detection}". In other words, we hypothesize that keyphrase generation is benefited from pre-training signals that help the model induce the key information from the context. 

Observing that keyphrases are often snippets or synonyms of salient in-text spans (which we call \textbf{salient spans}), we propose to derive learning signals from them for task-specific pre-training using PLMs. We posit that a span carries salient information if it can effectively identify the associated document. Based on this assumption, we design a retrieval-based salient span mining procedure that finds spans that are domain-wise salient and functionally similar to keyphrases. Using these spans, we design \textbf{salient span recovery (SSR)} and \textbf{salient span prediction (SSP)} as objectives to further pre-train BART \citep{lewis-etal-2020-bart} with unlabeled in-domain data. By corrupting salient spans from the document and asking the model to predict them back within or without the original context, SSR and SSP encourage the model to learn knowledge conducive to downstream keyphrase generation. 

We design low-resource benchmarks in the scientific domain and extensively compare our method with supervised and unsupervised keyphrase generation baselines. The results establish that the proposed method can outperform the BART fine-tuning baseline and various supervised keyphrase generation models trained from scratch in the low-resource setting. Moreover, we show that one variant of SSR is superior to other in-domain pre-training objectives, such as text infilling and title generation. Finally, we show that our method improves the performance of zero-shot domain transfer. We conclude by observing that manually annotated present keyphrases align with the assumptions of our retrieval-based span selection method. 

In summary, the main innovation of the paper is the strategy to select information from unlabeled data for effective learning of PLM-based low-resource keyphrase generation. We do not aim at designing masking strategies, as literature has explored closely related ones \citep{joshi-etal-2020-spanbert, guu2020realm}, or performing large-scale pre-training with \textit{annotated} keyphrase data, as explored in the concurrent work \citet{kulkarni2021learning}. Instead, we (1) observe that phrase saliency can be defined from the perspective of information retrieval, (2) design a procedure to mine salient spans automatically from large in-domain unlabeled data, (3) use these spans for domain-adaptive pre-training that teaches the model to induce essential information, and (4) demonstrate the resulting gains on low-resource keyphrase generation and zero-shot domain transfer. 
We release our experiment code and model outputs at \url{https://github.com/xiaowu0162/low-resource-kpgen} to facilitate future research.

\section{Methods}

\begin{figure*}[]
\includegraphics[width=\textwidth]{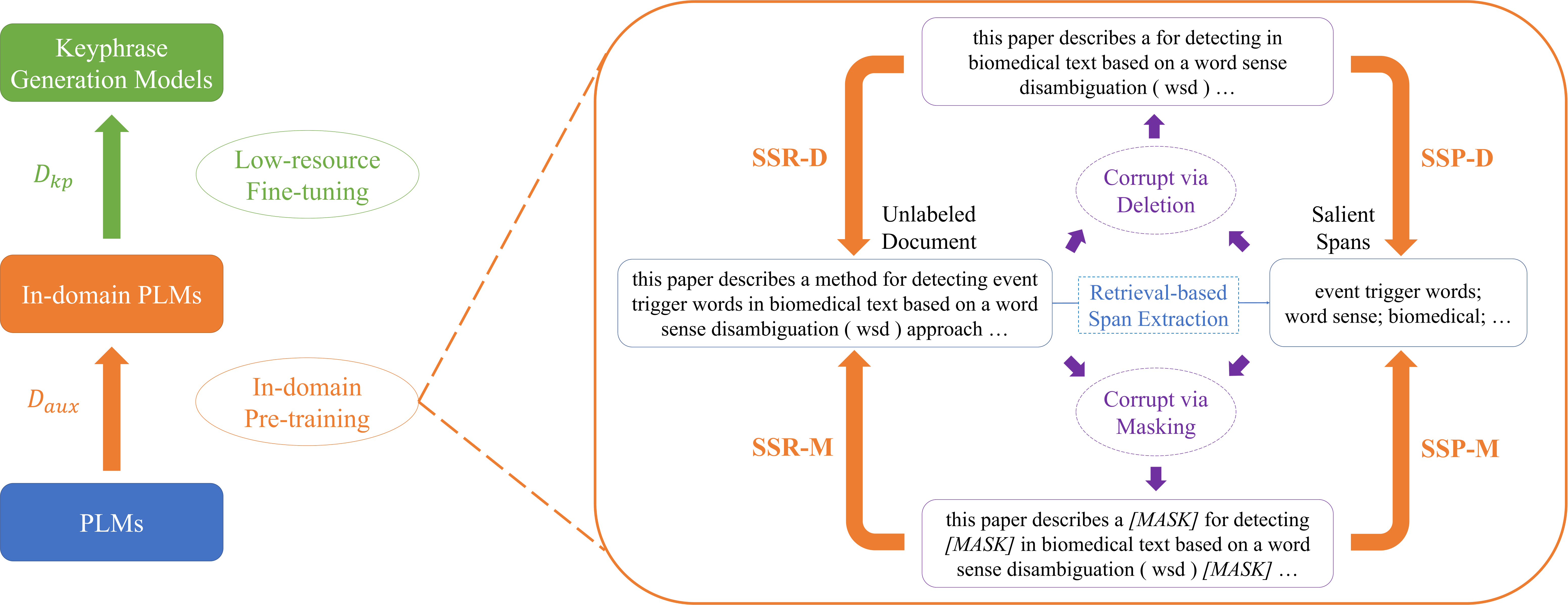}
\caption{An illustration of the proposed framework. A PLM is first pre-trained on large in-domain data $D_{aux}$ using one of the proposed objectives, and then fine-tuned on keyphrase generation using $D_{kp}$. In the example on the right, the salient span "event trigger words" and a random span "text" are corrupted, while "biomedical" is not. }
\label{masking-strategy}
\end{figure*}

\label{sec:methodology}
\paragraph{Problem Definition} 
Let $D_{kp}$ denote a keyphrase generation dataset, which is a set of tuples \textbf{$(\mathbf{x^i},\mathbf{p^i})$}, where $\mathbf{x^i}$ is an input document and $\mathbf{p^i}=\{p^i_1, p^i_2, ..., p^i_{|\mathbf{p^i}|}\}$ is the corresponding set of keyphrases (each of which is a sequence of tokens). Following \citet{yuan-etal-2020-one}, we define keyphrase generation as generating the sequence $\mathbf{y^i}=(p^i_1\ $\texttt{[sep]}$\ p^i_2\ $\texttt{[sep]}$\ ... \ $\texttt{[sep]}$\ p^i_{|\mathbf{p^i}|})$\footnote{We use semicolon as \texttt{[sep]} in our implementation.} based on the source text $\mathbf{x^i}$. In addition, let $D_{aux}$ be a set of \textit{unlabeled} documents from $D_{kp}$'s domain.

A typical way to train BART-like sequence-to-sequence PLMs for keyphrase generation is to directly fine-tune using the formulation above. Designed for small $D_{kp}$, our method first extracts salient spans using $D_{aux}$ and further trains BART using these spans. The resulting model with task-specific and domain-specific knowledge is then fine-tuned in the same way. Section \ref{subsec:retrieval-for-salient-span} introduces the salient span extraction method, and section \ref{subsec:in-domain-repr-learning} details the intermediate learning objectives.

\subsection{Retrieval for Salient Spans}
\label{subsec:retrieval-for-salient-span}
Inspired by previous works that identify retrieval as an important usage of keyphrases and a possible way to evaluate keyphrases \citep{kim-etal-2013-applying, boudin-etal-2020-keyphrase, boudin-gallina-2021-redefining}, we use retrieval as a tool to define and extract salient spans. Concretely, we define a salient span of a document as a contiguous sequence of tokens (an n-gram) that can retrieve the document from $D_{aux}$ via the BM25 retrieval \citep{10.1007/978-1-4471-2099-5_24}. For each document $\mathbf{x^i}\in D_{aux}$, let $Q^i=\{q^i_1, ..., q^i_n\}$ be a set of candidate n-grams. Let $BM25(x,q)$ be the BM25 score between a document $x$ and a query $q$ in $D_{aux}$. Then, define
\begin{multline*}
    rank(q^i_j)=|\mathbf{x'}\in D_{aux}: \\ BM25(\mathbf{x'},q^i_j)>BM25(\mathbf{x^i},q^i_j)|.
\end{multline*}
We then choose the set of salient spans $S^i$ from $Q^i$ by applying a filtering function to $rank(q^i_j)$.
\begin{multline*}
    S^i=\{q^i_j\in Q^i: rank(q^i_j)\leq threshold(|q^i_j|)\},
\end{multline*}
where $threshold(\cdot)$ is a function that specifies a maximum accepted $rank$ based on the span length. We use variable thresholds for different lengths to reduce BM25's bias towards longer phrases. 

Different from keyphrase extraction works that require the keyphrases to be noun phrases \citep{hulth-2003-improved, 10.5555/1620163.1620205, bougouin-etal-2013-topicrank}, we allow any n-gram from the document that does not contain stop words from being a candidate. To match the length of keyphrases, we require $n$ to be at most 3. In addition, different from previous works that use retrieval methods to identify similar documents and use their keyphrase annotations as external knowledge \citep{chen-etal-2019-integrated, kim-etal-2021-structure}, we use each candidate as a query and use the retrieved documents for calculating the $rank$.

Intuitively, our definition of salient spans reflects the idea that generating good keyphrases requires both \textit{intra-article} and \textit{inter-article} reasoning: while intra-article reasoning is used to find the most emphasized spans, inter-article knowledge is employed to determine whether a span can identify the article of interest in the sea of other articles. 

\subsection{In-domain Representation Learning}
\label{subsec:in-domain-repr-learning}
 After extracting the spans containing document-wise and domain-wise salient information, we propose to facilitate the downstream fine-tuning on the small $D_{kp}$ by first training BART on $D_{aux}$ with the following objectives.

\paragraph{Salient Span Recovery}
We design salient span recovery as a variant of BART's pre-training objectives where the tokens for masking or deletion are strategically chosen. Let $S^i=\{s^i_1, ..., s^i_n\}$ be the salient spans of $\mathbf{x^i}$. During training, each occurrence of $s^i_j$ in $\mathbf{x^i}$ is corrupted with probability $k_{s}$. In addition, we corrupt words in $\mathbf{x^i} \setminus (s^i_1\cup ...\cup s^i_n)$ randomly with probability $k_{o}$ to obtain the final input $\mathbf{x^i_{SSR}}$. The model is trained to minimize the cross entropy loss $\mathcal{L}_{CE}(\mathbf{z^i},\mathbf{x^i})$, where $\mathbf{z^i}$ is the model's reconstruction of the corrupted input $\mathbf{x^i_{SSR}}$. 

We experiment with two corruption strategies: (1) \textit{replacing} the salient spans or randomly selected words with a single \texttt{[MASK]} token in the input (denoted as \textbf{SSR-M}) or (2) \textit{deleting} the salient spans or randomly selected words from the input sequence (denoted as \textbf{SSR-D}).

\paragraph{Salient Span Prediction} We design SSP to align with the keyphrase generation task explicitly. While the input in SSP is still $\mathbf{x^i_{SSR}}$, the target is the concatenation of the salient spans $\mathbf{x^i_{SSP}}=(s^i_1\ $\texttt{[sep]}$\ s^i_2\ $\texttt{[sep]}$\ ... \ $\texttt{[sep]}$\ s^i_n)$, sorted by $rank(s^i_j)$ in the ascending order. The model is trained to minimize the cross entropy loss $\mathcal{L}_{CE}(\mathbf{z^i},\mathbf{x^i_{SSP}})$, where $\mathbf{z^i}$ is the model's prediction of the salient spans based on $\mathbf{x^i_{SSR}}$. 

Similar to SSR, we also experiment with two variants: \textbf{SSP-M} refers to \textit{replacing} the salient spans or randomly selected words with a single \texttt{[MASK]} token and \textbf{SSP-D} means \textit{deleting} the salient spans or randomly selected words from the input. Figure \ref{masking-strategy} demonstrates the four objectives. SSR-M uses the same input corruption strategy as SSP-M, and SSR-D uses the same input corruption strategy as SSP-D.

\section{Experimental Setup}

\subsection{Datasets}

We conduct evaluations on five scientific keyphrase generation datasets. We use KP20k \citep{meng-etal-2017-deep} for training and evaluate on KP20k, Inspec \citep{10.3115/1119355.1119383}, Krapivin \citep{Krapivin2009LargeDF}, NUS \citep{10.1007/978-3-540-77094-7_41}, and SemEval \citep{kim-etal-2010-semeval}. After removing articles overlapping with the validation or test set, the KP20k train set contains 509,818 instances. Following \citet{meng-etal-2017-deep}, we lower-case the text and replace the digits with a <digit> symbol to preprocess all the datasets. Table \ref{tab:test-sets-statistics} presents the statistics of the test datasets.

We use the KP20k train set to create $D_{kp}$ and $D_{aux}$, while keeping the validation and test sets the same. For the major results presented in section 4, we set $|D_{kp}|=20,000$ and we let $D_{aux}$ be the entire train set. In other words, only $20,000$ \textit{annotated} documents are available to the model. 

\subsection{Baselines}
First, we consider the following \textit{unsupervised} baselines. As most of these methods are keyphrase extraction methods except \citet{shen2021unsupervised}, we only evaluate their present keyphrase performance.

\textbf{TextRank} \citep{textrank2004} is a graph-based method that converts text to graphs and then uses PageRank to rank candidate phrases.

\textbf{SIFRank} and \textbf{SIFRank+} \citep{8954611} rank phrases by phrase-document cosine similarity with PLM-based dense embeddings. SIFRank+ uses position information to better handle long documents.

\textbf{\citet{liang-etal-2021-unsupervised}} is embedding-based and combines the global phrase-document similarity with the local boundary-aware degree centrality to calculate the score of each candidate phrase for ranking.

\textbf{AutoKeyGen} \citep{shen2021unsupervised} performs  keyphrase generation by constructing a phrase bank to predict present keyphrases via partial matching and to train a model to generate absent keyphrases.

We also consider the following \textit{supervised} baselines trained on the low-resource $D_{kp}$.

\textbf{ExHiRD-h} \citep{chen-etal-2021-training} designs a hierarchical decoding framework combined with a hard exclusion algorithm for reducing duplication, applied on the CatSeq models \citep{yuan-etal-2020-one}.

\textbf{One2Set} \citep{ye-etal-2021-one2set} proposes to train a transformer to predict keyphrases in parallel as a set based on learned control codes, which avoids the bias of generating keyphrases as a sequence.

\textbf{BART}. A fine-tuned BART-base \citep{lewis-etal-2020-bart} model for keyphrase generation.

\textbf{Transformer}. A randomly initialized Transformer with BART's architecture and vocabulary.

We denote our methods as \textbf{BART+SSR-M}, \textbf{BART+SSR-D}, \textbf{BART+SSP-M} and \textbf{BART+SSP-D}. They train BART on $D_{aux}$ using SSR or SSP, and then train on $D_{kp}$ for keyphrase generation.

\setlength{\tabcolsep}{3pt}
\begin{table}[!t]
    \centering
    {%
    \begin{tabular}{l | c  c  c c}
    \hline
    Dataset & \#Examples & \#KP & |KP| & $\%$AKP \\
    \hline
    KP20k & 20,000 & 5.28 & 2.04 & 37.06 \\
    Inspec & 500 & 9.83 & 2.48 & 26.38\\
    Krapivin & 400 & 5.85 & 2.21 & 44.34\\
    NUS & 211 & 11.65 & 2.22 & 45.61\\
    SemEval & 100 & 14.66 & 2.38 & 57.37\\
    \hline
    \end{tabular}
    }
    \caption{Statistics of all the test sets we use. \#KP: average number of keyphrases of each document; |KP|: average length of each keyphrase; $\%$AKP: the percentage of absent keyphrases.}
    \label{tab:test-sets-statistics}
\end{table}

\subsection{Evaluation}
Following \citet{chan-etal-2019-neural}, we use greedy decoding. We apply the Porter Stemmer \citep{Porter1980AnAF} on the predictions and targets and then calculate the macro-averaged F1@5 and F1@M for present and absent keyphrases. While F1@$k$ only considers the top $k$ predictions for evaluation, F1@M takes all predictions from the model \citep{yuan-etal-2020-one}. We do not calculate F1@M for the unsupervised methods since they only predict the ranking of the candidates. Each experiment is repeated with three randomly sampled $D_{kp}$'s, and we report the averaged scores. Unless otherwise stated, we use the same script based on \citet{chan-etal-2019-neural}'s implementation to calculate the scores.

\subsection{Implementation Details}
In this section, we provide the implementation details. Further discussions on the baselines and the hyperparameters are provided in the appendix.

\paragraph{SSR and SSP} We obtain the salient spans via BM25 retrieval. Using Elasticsearch\footnote{\url{https://github.com/elastic/elasticsearch}}, we build a database containing documents from $D_{aux}$. Then, for each document in $D_{aux}$, we construct a boolean query to perform a fuzzy search for each of its candidates. We use BM25 as the search metric, using $k_1=1.2$ and $b=0.75$. Our query code is based on the implementations of \citet{xorqa}. Then, we use the following $threshold$ function:
$$threshold=\{1: 500, 2: 430, 3: 360\}$$

We start training from the pre-trained BART-base checkpoint using Fairseq's \texttt{translation} task\footnote{\url{https://github.com/facebookresearch/fairseq}}. The input documents are truncated to 512 tokens. We set $k_s=0.4$ and $k_o=0.2$. This gives a corruption rate of about $39\%$ tokens, and the \texttt{[MASK]} symbol takes up about $11\%$ of the resulting corrupted text (for SSR-M and SSP-M).  For SSP-M and SSP-D, we remove phrases from the target that are substrings of longer salient spans. We use the Adam optimizer with $\beta_1 = 0.9$, $\beta_2 = 0.999$ and polynomial decay with 6000 warmup steps. We use batch size 64, learning rate 3e-4, 0.1 dropout, and 0.01 weight decay.

\paragraph{Fine-tuning} For fine-tuning on $D_{kp}$, we use learning rate 1e-5, batch size 32, and 150 warmup steps. All experiments are run on two Nvidia GTX 1080Ti GPUs, and we use gradient accumulation to achieve the desired batch size.

\section{Results and Analysis}

\setlength{\tabcolsep}{4pt}
\begin{table*}[h]
    \centering
    \begin{tabular}{l | c c | c c | c c | c c | c c }
    \hline
    \multirow{2}{*}{Method} & \multicolumn{2}{c|}{\textbf{KP20k}} & \multicolumn{2}{c|}{\textbf{Inspec}} & \multicolumn{2}{c|}{\textbf{Krapivin}} & \multicolumn{2}{c|}{\textbf{NUS}} & \multicolumn{2}{c}{\textbf{SemEval}} \\
    & F1@5 & F1@M & F1@5 & F1@M & F1@5 & F1@M & F1@5 & F1@M & F1@5 & F1@M \\
    \hline
    ExHiRD-h & 0.35 & 0.57 & 0.26 & 0.41 & 0.65 & 0.98 & 0.46 & 0.57 & 0.43 & 0.56 \\
    One2Set & 0.54 & 0.98 & 0.10 & 0.15 & 0.71 & 1.32 & 0.69 & 1.01 & 0.66 & 0.94 \\
    Transformer & 1.16 & 1.90 & 0.48 & 0.71 & 1.30 & 1.86 & 1.50 & 2.02 & 1.17 & 1.44 \\
    \hline
    BART & 0.93 & 1.87 & 0.89 & 1.58 & 1.37 & 2.52 & 1.06 & 1.70 & 0.87 & 1.24 \\
    BART+SSP-M & 1.39 & 2.78 & 0.93 & 1.70 & 2.24 & 4.34 & 1.77 & 2.92 & 1.66 & 2.31 \\
    BART+SSP-D & 1.35 & 2.73 & 0.91 & 1.63 & 2.19 & 4.06 & 1.86 & 2.79 & 1.28 & 1.78 \\
    BART+SSR-M & \textbf{1.95} & 3.42 & 1.04 & 1.73 & 2.41 & 3.87 & 2.16 & 3.12 & 1.85 & 2.39 \\
    BART+SSR-D & \textbf{1.95} & \textbf{3.76} & \textbf{1.22} & \textbf{2.07} & \textbf{2.55} & \textbf{4.63} & \textbf{3.11} & \textbf{5.31} & \textbf{2.15} & \textbf{2.89} \\
    \hline 
    \end{tabular}
    \caption{F1 scores of \textit{low-resource} absent keyphrase generation on five scientific benchmarks ($|D_{kp}|$=20,000). Best result is boldfaced. BART+SSR-D outperforms the other approaches in all benchmarks. Meanwhile, all the proposed objectives improve over simple BART fine-tuning. }
    \label{tab:scikp-main-results}
\end{table*}
\setlength{\tabcolsep}{3.5pt}
\begin{table*}[!t]
    \centering
    \begin{tabular}{l | c c | c c | c c | c c | c c }
    \hline
    \multirow{2}{*}{Method} & \multicolumn{2}{c|}{\textbf{KP20k}} & \multicolumn{2}{c|}{\textbf{Inspec}} & \multicolumn{2}{c|}{\textbf{Krapivin}} & \multicolumn{2}{c|}{\textbf{NUS}} & \multicolumn{2}{c}{\textbf{SemEval}} \\
    & F1@5 & F1@M & F1@5 & F1@M & F1@5 & F1@M & F1@5 & F1@M & F1@5 & F1@M \\
    \hline
    TextRank & 9.24 & - & 32.07 & - & 11.56 & - & 8.99 & - & 9.24 & - \\
    SIFRank  & 14.09 & - & \textbf{38.22} & - & 15.94 & - & 13.97 & - & 16.43 & - \\
    SIFRank+  & 20.00 & - & 35.08 & - & 19.59 & - & 25.47 & - & 24.77 & - \\
    AutoKeyGen & 23.4 & - & 30.3 & - & 17.1 & - & 21.8 & - & 18.7 & - \\
    \citet{liang-etal-2021-unsupervised}  & 17.66 & - & 29.57 & - & 16.93 & - & 24.98 & - & 25.33 &  -\\
    \hline
    ExHiRD-h & 24.01 & 29.92 & 22.41 & 25.21 & 22.83 & 29.32 & 28.26 & 33.75 & 22.23 & 26.71 \\
    One2Set & 15.76 & 23.84 & 10.46 & 14.21 & 15.23 & 23.24 & 20.61 & 28.22 & 15.11 & 20.48 \\
    Transformer & 11.06 & 18.04 & 6.63 & 9.91 & 10.05 & 17.12 & 14.51 & 20.72 & 8.77 & 12.13 \\
    \hline
    BART & 26.97 & 31.54 & 28.54 & \textbf{33.93} & 26.62 & 31.12 & 33.88 & 38.08 & \textbf{26.33} & \textbf{30.12} \\
    BART+SSP-M & 28.04 & 32.30 & 27.39 & 32.25 & \textbf{27.51} & 33.59 & \textbf{34.35} & 39.21 & 24.49 & 27.72 \\
    BART+SSP-D & 28.29 & 32.63 & 27.29 & 32.84 & 27.46 & 32.49 & 33.44 & 38.05 & 26.04 & 29.47 \\
    BART+SSR-M & 25.83 & 33.00 & 22.57 & 28.09 & 23.18 & 30.01 & 31.13 & 36.86 & 22.60 & 27.28 \\
    BART+SSR-D & \textbf{28.82} & \textbf{35.43} & 24.35 & 30.17 & 27.08 & \textbf{34.30} & 34.34 & \textbf{40.49} & 23.69 & 29.04 \\
    \hline
    \end{tabular}
    \caption{F1 scores of \textit{low-resource} present keyphrase generation on five benchmarks in the scientific domain ($|D_{kp}|$=20,000). Best result is boldfaced. Pre-trained language models greatly outperform methods trained from scratch. Moreover, performing in-domain pre-training using the proposed objectives improves over the simple BART fine-tuning on three of the five benchmarks. Some example outputs are presented in the appendix.}
    \label{tab:scikp-main-results-pkp}
\end{table*}

We aim to address the following questions.
\begin{compactenum}
    \item Does our method learn strong representations from unlabeled data, and thus has competitive performance in low-resource fine-tuning?
    \item Can our method outperform training on $D_{aux}$ with other objectives such as text infilling?
    \item Can our representations benefit keyphrase generation in zero-shot settings?
    \item Are present keyphrases effective for retrieval? How much do they overlap with salient spans?
\end{compactenum}

\subsection{Low-Resource Fine-tuning Performance}

\begin{figure}[]
\centering
\input{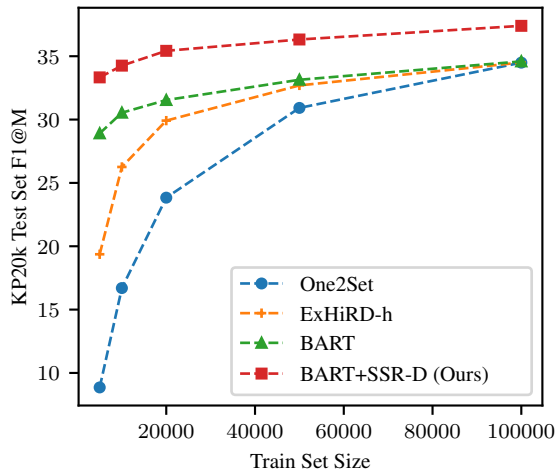}
\caption{Present keyphrase generation performance of different methods as a function of train set size.  Our in-domain annotation-free pre-training approach achieves the best performance in all resource schemes.}
\label{perf-vs-resource}
\end{figure}

The scarcity of annotated data poses a significant challenge to supervised keyphrase generation models. Using $D_{kp}$ from KP20k with size 5k, 10k, 20k, 50k, and 100k, we train One2Set \citep{ye-etal-2021-one2set} and ExHiRD-h \citep{chen-etal-2021-training} from scratch and compare their performance with fine-tuning the pre-trained BART or our BART+SSR-D model trained on KP20k. The macro-averaged F1@M scores for the present keyphrases of the KP20k test set are shown in Figure \ref{perf-vs-resource}. One2Set and ExHiRD-h perform poorly with less than 50k training data and have a similar performance as BART when the data size is as large as 100k. Nevertheless, in all resource regimes, our in-domain unsupervised SSR-D pre-training achieves
the best performance.

Next, we focus on the scenario with $|D_{kp}|$ = 20,000 and provide a more careful analysis. Table \ref{tab:scikp-main-results} and \ref{tab:scikp-main-results-pkp} show the performance of low-resource absent and present keyphrase generation on the scientific benchmarks. Additional qualitative results are presented in the appendix.

\paragraph{Using pre-trained language models improves low-resource present keyphrase performance.} From Table \ref{tab:scikp-main-results-pkp}, it is apparent that fine-tuning BART significantly outperforms the three supervised baselines trained from scratch. However, Table \ref{tab:scikp-main-results} indicates that the absent keyphrase generation follows a different pattern. Randomly initializing a Transformer with BART's architecture, we achieve better F1@5 and F1@M on KP20k, NUS, and SemEval.  This shows that in the low-resource regime, BART pre-training mainly facilitates present keyphrase generation but does not give the model much additional capability to generate absent keyphrases.

\paragraph{SSR-D performs the best in the proposed objectives.} Among the proposed objectives, we find that SSR-D enables the best fine-tuning performance, achieving the best F1@5 and F1@M for absent keyphrase generation on all datasets and the best F1@M for present keyphrase generation on three of the five datasets. Our intuition is that SSR-D is the most challenging objective because it requires the prediction of target spans at the correct positions in the context (rather than only predicting the salient spans in any order as in SSP), without being given \texttt{[MASK]} tokens as hints in the input (as in SSR-M or SSP-M). SSR-D's low-resource absent keyphrase performance is highly competitive. Its F1 scores on KP20k, Inspec, Krapivin, and SemEval even exceed those of ExHiRD-h trained on the complete KP20k train set (as reported in \citet{chen-etal-2021-training} and Table \ref{tab:full-train-results} in the appendix).

At the same time, we find SSP-M and SSP-D have very similar performance, while SSR-D outperforms both on KP20k, Krapivin, and NUS. One possible reason is that they converge in a relatively short time, and thus the behaviors do not differ a lot. Also, they may be affected by the noise in the salient spans due to the lack of human annotation. We suspect that SSP-like objectives may have more advantages if the span quality is as good as manual annotations, as suggested by the observations made by \citet{kulkarni2021learning}.

\subsection{In-domain Pre-training Objectives}
In this section, we compare SSP and SSR with two baseline objectives that can be used to train BART on $D_{aux}$ before fine-tuning on $D_{kp}$. 

\paragraph{BART+TI} \emph{Text infilling} (TI) is one of the pre-training objectives of BART. In text infilling, spans with lengths following a Poisson distribution ($\lambda=3$) are randomly selected from $\mathbf{x^i}$ and replaced with a single \texttt{[MASK]} token to obtain $\mathbf{x^i_{Infilling}}$. The model is trained to minimize the cross entropy loss $\mathcal{L}_{CE}(\mathbf{z^i},\mathbf{x^i})$, where $\mathbf{z^i}$ is the model's reconstruction of the corrupted input $\mathbf{x^i_{Infilling}}$.

\paragraph{BART+TG} \citet{ye-wang-2018-semi} showed that learning signals from \emph{title generation} can benefit low-resource keyphrase generation. We remove the titles from $\mathbf{x^i}$ and further pre-train BART for generating the titles using cross-entropy loss.

\begin{figure}[]
\centering
\input{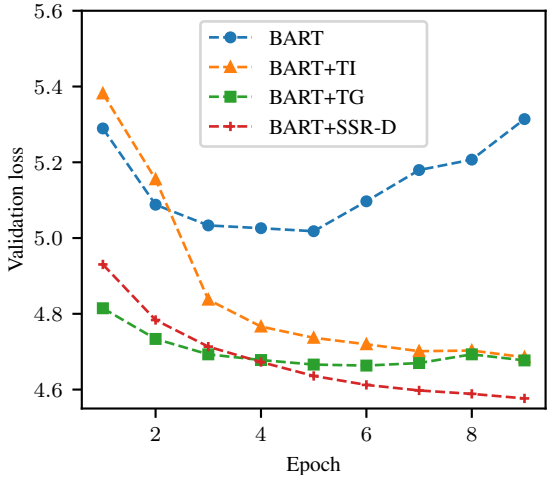}
\caption{KP20k fine-tuning validation loss with different initializations, using learning rate 1e-5 and $|D_{kp}|=20,000$. BART+SSR-D converges to the lowest loss and suffers the least from overfitting.}
\label{loss-comparison-kp20k}
\end{figure}

\begin{table}[]
    \centering
    \begin{tabular}{l | c c | c c}
    \hline
    \multirow{2}{*}{Method} & \multicolumn{2}{c|}{\textbf{Present}} & \multicolumn{2}{c}{\textbf{Absent}} \\
    & F1@5 & F1@M & F1@5 & F1@M  \\
    \hline
    BART+TI & 29.27 & 34.58 & 1.33 & 2.67\\
    BART+TG & \textbf{29.77} & 33.86 & 1.28 & 2.55\\
    BART+SSR-D & 28.82 & \textbf{35.43} & \textbf{1.95} & \textbf{3.76} \\
    \hline
    \end{tabular}
    \caption{F1 scores of low-resource keyphrase generation on KP20k ($|D_{kp}|$=20,000) based on in-domain models pre-trained with different methods. BART+SSR-D achieves the best F1@5 and F1@M for absent keyphrases, and the best F1@M for present keyphrases.}
    \label{tab:in-domain-comparison}
\end{table}

\paragraph{Results} Table \ref{tab:in-domain-comparison} compares BART+TI, BART+TG, and BART+SSR-D. Fine-tuning via BART+SSR-D achieves the best F1@5 and F1@M for absent keyphrases and the best F1@M for present keyphrases. This indicates that SSR is more tailored for identifying keyphrases than TI. Also, TG contributes better to present keyphrases since the information in titles is likely to be extractive. 

In Figure \ref{loss-comparison-kp20k}, we plot the validation loss for low-resource fine-tuning. We observe that all in-domain pre-training methods outperform the BART fine-tuning baseline. Initializing with BART+SSR-D converges to the best validation loss and seems less susceptible to overfitting on the small data.

\subsection{Zero-shot Cross-domain Generalization}

Although we mainly focus on the low-resource scheme, it is also helpful to investigate the zero-shot generalization ability. Using the in-domain models trained with KP20k as $D_{aux}$, we fine-tune the models on keyphrase generation using KPTimes \citep{gallina-etal-2019-kptimes} and evaluate on the scientific benchmarks. In this setting, while KPTimes provides comprehensive task-wise information, the final performance also highly depends on how much domain-wise information the model extracts from $D_{aux}$.

We compare the performance of BART+TI, BART+TG, and BART+SSR-D. The results are presented in Table \ref{tab:kp20k-zeroshot-results}. Although the intermediate training does not use manual keyphrase labels, the learned representation condenses domain-specific knowledge. It results in better zero-shot transfer performance compared to the BART directly fine-tuned on KPTimes. SSR-D achieves the best cross-domain transfer performance, outperforming the other methods by a large margin, especially in present keyphrase generation and F1@5 for absent keyphrase generation. We also directly report the score of the intermediate SSP-D model. Despite a somewhat competitive performance on present keyphrases, its absent keyphrase performance is worse than the baselines. Considering the poor performance of BART fine-tuned on KPTimes, we conclude that training with in-domain annotated data is crucial for absent keyphrase generation.

\begin{table}[]
    \centering
    \resizebox{\linewidth}{!}
    {%
    \begin{tabular}{l | c c | c c }
    \hline
    \multirow{2}{*}{Method} & \multicolumn{2}{c|}{\textbf{Present}} & \multicolumn{2}{c}{\textbf{Absent}} \\
    & F1@5 & F1@M & F1@5 & F1@M\\
    \hline
    SSP-D-only & 4.21 & 5.63 & 0.08 & 0.11 \\
    \hline
    BART & 3.01 & 5.51 & 0.13 & 0.23 \\
    BART+TI & 6.51 & 11.13 & 0.22 & 0.40 \\
    BART+TG & 7.20 & 12.37 & 0.27 & \textbf{0.50} \\
    BART+SSR-D & \textbf{10.81} & \textbf{16.87} & \textbf{0.82} & 0.47 \\
    \hline
    \end{tabular}
    }
    \caption{F1 scores of zero-shot keyphrase generation on KP20k. Best result is boldfaced. "SSP-D-only" = BART trained on SSP-D using KP20k. BART+SSR-D significantly outperforms other methods.}
    \label{tab:kp20k-zeroshot-results}
\end{table}

\subsection{Analysis of BM25 Retrieval}
In this section, we address questions about our retrieval-based definition of salient spans.

\paragraph{Can present keyphrases retrieve well?} 
We construct a document pool with the train and validation set of KP20k and the five test datasets. For each document, we perform BM25 retrieval using each of its present keyphrases. If the document is retrieved in the top 1000 documents, then we consider the retrieval as successful. Table \ref{tab:bm25-pkp-success-rate} presents the resulting success rates. SemEval is excluded because all of its keyphrases are stemmed. We observe that the overall success rate is high for all datasets. This shows that the properties of present keyphrases align with our retrieval-based definition of salient spans. Moreover, shorter keyphrases retrieve worse due to their higher frequency in the corpus. This justifies our design of the length-adaptive \textit{threshold} function to compensate for the bias.

\paragraph{How do present keyphrases overlap with salient spans?}
We compute the overlap between the salient spans and the actual present keyphrases. For each document, we define \textbf{phrase recall} as the proportion of present keyphrases that are present in the salient span, \textbf{word recall} as the proportion of all words in present keyphrases that are also in any salient span, and \textbf{word precision} as the proportion of words in salient spans that are included in any keyphrase of the same document. Table \ref{tab:bm25-overlap-eval} presents the measures evaluated on the KP20k train set. The columns labeled "len k" only consider keyphrases and salient spans of length k. We observe that the salient spans can cover about 36\% of the present keyphrases and 85\% of the words in the present keyphrases.
Meanwhile, the 13\% precision indicates that the salient spans also contain many words that do not belong to any keyphrase. In addition, although we tune the $threshold$ function to benefit short phrases, the overlap between the salient single-word spans and the present single-word keyphrases is still small. Also, the overall word precision is much higher than obtained by considering the phrase lengths separately. This suggests that our method tends to ignore the boundaries of keyphrases. 

\begin{table}[!t]
    \centering
    \begin{tabular}{l | c c c c}
    \hline
    Dataset & len 1 & len 2 & len 3 & overall \\
    \hline
    \textbf{KP20k} & 39.4\% & 83.5\% & 91.7\% & 80.5\% \\
    \textbf{Inspec} & 67.8\% & 89.9\% & 97.8\% & 90.4\% \\
    \textbf{Krapivin} & 52.2\% & 82.7\% & 94.9\% & 81.1\% \\
    \textbf{NUS} & 52.4\% & 77.5\% & 93.7\% & 76.1\% \\
    \hline
    \end{tabular}
    \caption{Retrieval success rates of manually annotated present keyphrases. The success rate is high for all the datasets overall, while at the same time exhibiting a positive correlation with keyphrase length. }
    \label{tab:bm25-pkp-success-rate}
\end{table}

\begin{table}[!t]
    \centering
    \resizebox{\linewidth}{!}
    {
    \begin{tabular}{l | c c c c}
    \hline
    Measure & len 1 & len 2 & len 3 & overall \\
    \hline
    \textbf{Phrase Recall} & 0.188 & 0.376 & 0.380 & 0.364 \\
    \textbf{Word Recall} & 0.376 & 0.857 & 0.864 & 0.849 \\
    \textbf{Word Precision} & 0.039 & 0.069 & 0.051 & 0.128 \\
    \hline
    \end{tabular}
    }
    \caption{Overlap between salient spans and the present keyphrases of KP20k training set. Salient spans obtained using BM25 has high word-level coverage but lower phrase-level coverage. }
    \label{tab:bm25-overlap-eval}
\end{table}

\section{Related Work}

\paragraph{Low-resource Keyphrase Generation} 
Prior works in keyphrase identification are broadly divided into keyphrase extraction and keyphrase generation. 
While keyphrase extraction only extracts present keyphrases as spans of the document \citep{hulth-2003-improved,mihalcea-tarau-2004-textrank,10.5555/1620163.1620205,bougouin-etal-2013-topicrank,zhang-etal-2016-keyphrase,8954611, liang-etal-2021-unsupervised}, keyphrase generation directly predicts both present and absent keyphrases \citep{meng-etal-2017-deep, chen-etal-2018-keyphrase, Chen2019TitleGuidedEF, zhao-zhang-2019-incorporating, chan-etal-2019-neural, yuan-etal-2020-one, swaminathan-etal-2020-preliminary, ahmad-etal-2021-select, ye-etal-2021-one2set,kim-etal-2021-structure}. One solution to the "low-resource" problem is unsupervised keyphrase extraction or generation, which does not require annotations. However, they either cannot predict absent keyphrases or require the construction of large phrase banks and may have inferior performance compared to supervised methods. Alternatively, other previous studies have considered solving low-resource keyphrase generation via synthetic labeling and semi-supervised multi-task learning to leverage $D_{aux}$ \citep{ye-wang-2018-semi} or using reinforcement learning to exploit learning signals from a pre-trained discriminator in the setting of Generative Adversarial Networks \citep{lancioni-etal-2020-keyphrase}. In contrast, our innovation is the retrieval-based task-specific pre-training of PLMs.

\paragraph{Retrieval-Augmented Keyphrase Generation} 
Retrieval methods have been used to investigate keyphrases' role or to enhance the performance of keyphrase generation models. \citet{kim-etal-2013-applying} and \citet{boudin-etal-2020-keyphrase} verify that keyphrases can significantly enhance retrieval performance. \citet{boudin-gallina-2021-redefining} provide a finer-grained analysis of absent keyphrases and conclude that a subset of them contributes to information retrieval by adding in new information via document expansion. \citet{chen-etal-2019-integrated} design a retriever to find similar documents from the training corpus, whose phrases are used as keyphrase candidates and encoded as a continuous vector to augment the input. \citet{kim-etal-2021-structure} propose to augment the document's structure with keyphrases from similar documents and obtain a structure-aware representation of the augmented text. 

\paragraph{Language modeling and keyphrase generation} 
Recent studies have successfully used PLMs for rich-resource keyphrase generation \citep{liu2020keyphrase} and keyphrase extraction \citep{sahrawat2019keyphrase}. For other tasks, studies explored continued domain-adaptive pre-training of the autoencoding \citep{gururangan-etal-2020-dont, DBLP:journals/corr/abs-1901-08746} and encoder-decoder PLMs \citep{yu-etal-2021-adaptsum}. 
\citet{kulkarni2021learning} is a concurrent work that explores a similar objective for representation learning using supervised data.
In comparison, our work focuses on unsupervised learning to facilitate low-resource keyphrase generation. It thus leads to different conclusions from that in \citet{kulkarni2021learning}. 



\section{Conclusion}
This paper considers the problem of low-resource keyphrase generation. We design an innovative retrieval-based method to extract salient information from unlabeled documents and perform continued BART pre-training. We verify that the method facilitates low-resource keyphrase generation and zero-shot cross-domain generalization. Our method consistently outperforms the baselines in a range of resource schemes. Future works may consider investigating dense embeddings for extracting salient spans, composing the proposed objectives, or designing specialized methods for fine-tuning on small datasets. 

\section*{Limitations}
In this work, although we conduct experiments in a variety of settings and on several datasets, most of them are only in the scientific domain. In addition, we only experiment on BART. We use BART because it is pre-trained using denoising autoencoding, which is closer to salient span recovery and prediction than other PLMs such as T5 \citep{raffel2020exploring}. Finally, we acknowledge that the proposed large-scale intermediate representation learning causes energy costs and emissions. As a trade-off, we obtain strong representations to solve the challenging low-resource problem better and to be reused for fine-tuning on different datasets. 

\section*{Ethical Statement}
We use the KP20k dataset distributed by their original host, and we have verified that our preprocessing methods do not introduce external biases or sensitive information. However, our self-supervised representation learning method may propagate the bias that lies in the unlabeled external data it uses. As our approach can be easily integrated into BART-based keyphrase generation services, we encourage potential users to monitor for biases closely and apply corresponding mitigation measures when necessary.

\section*{Acknowledgment}
The research is supported in part by Taboola and an Amazon AWS credit award. We thank the Taboola team for helpful discussions and feedback. We also thank the anonymous reviewers and the members of UCLA-NLP for providing their valuable feedback.

\bibliographystyle{acl_natbib}
\bibliography{anthology,custom}

\clearpage
\appendix

\twocolumn[{%
 \centering
 \Large\bf Supplementary Material: Appendices \\ [20pt]
}]

\section{Rich-resource Results}
\begin{table*}[]
    \centering
    \begin{tabular}{l | c c | c c | c c | c c | c c }
    \hline
    \multirow{2}{*}{Method} & \multicolumn{2}{c|}{\textbf{KP20k}} & \multicolumn{2}{c|}{\textbf{Inspec}} & \multicolumn{2}{c|}{\textbf{Krapivin}} & \multicolumn{2}{c|}{\textbf{NUS}} & \multicolumn{2}{c}{\textbf{SemEval}} \\
    & F1@5 & F1@M & F1@5 & F1@M & F1@5 & F1@M & F1@5 & F1@M & F1@5 & F1@M \\
    \hline
    \multicolumn{11}{l}{\textbf{Present Keyphrase Generation}} \\
    \hline
    ExHiRD-h & 31.07 & 37.38 & 25.35 & 29.13 & 28.56 & 30.75 & - & - & 30.40 & 28.21 \\
    ExHiRD-s & 30.75 & 37.20 & 23.53 & 27.81 & 27.84 & 33.84 & - & - & 26.71 & 31.41 \\
    One2Set & \textbf{35.57} & \textbf{39.14} & \textbf{29.13} & 32.77 & \textbf{33.46} & \textbf{37.47} & \textbf{39.94} & \textbf{44.58} & \textbf{32.17} & 34.18 \\
    BART & 32.21 & 39.03 & 27.31 & \textbf{33.01} & 26.42 & 33.11 & 36.66 & 43.09 & 28.32 & \textbf{34.53} \\
    \hline
    \multicolumn{11}{l}{\textbf{Absent Keyphrase Generation}} \\
    \hline
    ExHiRD-h & 1.57 & 2.47 & 1.09 & 1.64 & 2.19 & 3.31 & - & - & 1.58 & 2.05 \\
    ExHiRD-s & 1.36 & 2.22 & 0.95 & 1.56 & 1.63 & 2.59 & - & - & 1.24 & 1.87 \\
    One2Set & \textbf{3.54} & \textbf{5.82} & \textbf{1.91} & \textbf{2.99} & \textbf{4.49} & \textbf{7.16} & \textbf{3.74} & \textbf{5.52} & \textbf{2.24} & \textbf{2.87} \\
    BART & 2.06 & 3.96 & 0.86 & 1.53 & 2.82 & 4.95 & 2.52 & 4.14 & 1.50 & 2.03 \\
    \hline
    \end{tabular}
    \caption{Rich-resource keyphrase generation results. All the scores reported are macro-averaged f1 scores across runs with three different seeds. Best result is boldfaced. We run our evaluation script on the predictions provided by \citet{chen-etal-2021-training} to get the scores for ExHiRD-h and ExHiRD-s. Although BART does not have SOTA performance, it is a competitive model for both present and absent keyphrase generation.}
    \label{tab:full-train-results}
\end{table*}

In Table \ref{tab:full-train-results}, we compare BART, ExHiRD-s \citep{chen-etal-2021-training}, ExHiRD-h \citep{chen-etal-2021-training}, and One2Set \citep{ye-etal-2021-one2set} trained on the entire KP20k train set. In this rich-resource scenario, BART fine-tuning outperforms ExHiRD on the scientific benchmarks while performing worse than One2Set on most datasets.

\section{Hyperparameter Optimization}
For SSP and SSR, we search over $\{\{1: 500, 2: 430, 3: 360\}, \{1: 500, 2: 400, 3: 300\}, \{1: 300, 2: 300, 3: 300\}\}$ for $threshold$, $\{0.3, 0.35, 0.4, 0.45\}$ for $k_s$ and $\{0.2, 0.3, 0.4\}$ for $k_o$. We also search over \{3e-4, 1e-4, 3e-5\} for learning rate. We prepare the validation set using the same method for each experiment and use validation loss as the stopping criteria during training. We choose the hyperparameters that enable the best validation performance during downstream fine-tuning on $D_{kp}$. 

For fine-tuning, we perform a grid search over \{1e-4, 6e-5, 3e-5, 1e-5\} for learning rate, \{32, 64\} for batch size, and \{50, 150, 400, 1000\} for the number of warmup steps. We choose the hyperparameters based on validation performance.
 
In Table \ref{tab:hyperparams}, we present all the hyperparameters for training our SSR/SSP model and fine-tuning on low-resource keyphrase generation. 
\begin{table*}[h]
    \centering
    \begin{tabular}{l | c | c | c }
    \hline
    Parameter & SSR & SSP & Fine-tuning \\
    \hline
    vocabulary size & 51,200 & 51,200 & 51,200 \\
    \# parameters & 140M & 140M & 140M \\
    $k_s, k_o$ & 0.4, 0.2 & 0.4, 0.2 & - \\
    total epochs & 60 & 10 & 15 \\
    batch size & 64 & 64 & 32 \\
    learning rate & 3e-4 & 3e-4 & 1e-5 \\
    lr schedule & polynomial & polynomial & polynomial \\
    warmup steps & 6000 & 6000 & 150\\
    optimizer & Adam & Adam & Adam\\
    weight decay & $0.01$ & $0.01$ & $0.01$ \\
    dropout & $0.1$ & $0.1$ & $0.1$ \\
    max. grad. norm & 0.1 & 0.1 & 0.1 \\
    \hline
    \end{tabular}
    \caption{Hyperparameters for pre-training using SSR or SSP and fine-tuning on low-resource keyphrase generation. "polynomial" means the \hyperlink{https://github.com/facebookresearch/fairseq/blob/main/fairseq/optim/lr_scheduler/polynomial_decay_schedule.py}{polynomial decay learning rate schedule}. "max. grad. norm" means the maximum norm allowed for the gradient. }
    \label{tab:hyperparams}
\end{table*}

\section{Implementation Details of the Baselines}
We use the publicly available implementations to reproduce \href{https://github.com/Chen-Wang-CUHK/ExHiRD-DKG}{ExHiRD-h}, \href{https://github.com/jiacheng-ye/kg_one2set}{One2Set}, \href{https://github.com/boudinfl/pke}{TextRank},
\href{https://github.com/sunyilgdx/SIFRank}{SIFRank}, and \href{https://github.com/sunyilgdx/SIFRank}{SIFRank+}. We use the scores reported by the authors for AutoKeyGen. For ExHiRD-h, we use the hyperparameters recommended in \citet{chen-etal-2021-training}. For One2Set, we use the recommended hyperparameters in the authors' implementation, except for removing dropout after tuning on the KP20k validation set. For SIFRank and SIFRank+, we use the L1 layer of ElMo, and set $\lambda=0.6$. We write our own implementations for \citet{liang-etal-2021-unsupervised}, where we follow the methods in SIFRank to generate candidate phrases and use \href{https://huggingface.co/docs/transformers/model_doc/bert}{BERT-base-uncased} \citep{devlin-etal-2019-bert} to obtain the contextual embeddings. Through a hyperparameter search on the KP20k validation set, we determine the set of hyperparameters $\{\alpha=1.2, \beta=0.0, \lambda=0.8\}$. 


\section{Characteristics of Salient Spans}
\paragraph{How many salient spans do we get?} In our BM25 retrieval setting, where the KP20k train set is used as $D_{aux}$, several spans can accurately retrieve the original document. On average, each document has 9.83 spans that can retrieve the document back to the top. Among these spans, 12\% are unigrams, 30\% are bigrams, and 58\% are trigrams. If exact matching is specified in Elasticsearch, the number of hits of the salient spans is low, indicating that they tend to be rare.

\paragraph{Is BM25 indispensable?} We considered the TF-IDF score \citep{tf-idf} as an alternative phrase-document similarity measure. We observed that it also gives good salient span predictions when the document lengths are similar. On KP20k, we find that using the \textit{retrieval scheme} is more important than choosing between TF-IDF and BM25 as the scoring function. We finally chose BM25 because it is designed for information retrieval, can better adapt to long documents, and enables better keyphrase generation performance. It is worth noting that it might help improve the scoring function by considering dense embeddings. We leave this to future work. 

\section{Further Discussions}

\paragraph{Failed Attempts}
To explore the possibility of extending or unifying our proposed objectives, we ran several preliminary experiments on (1) combining span masking with span deletion and (2) combining SSR and SSP via multi-task learning or multi-step adaptation. However, the results were not as good as BART+SSR-D.

\paragraph{Computational Budget}
All experiments are run on a local GPU server. SSP and SSR take 20 and 120 GPU hours, respectively, on a dataset of a size similar to KP20k, and the final fine-tuning takes 1 GPU hour on a dataset with 20,000 examples. 

\section{Qualitative Results}
We present two sets of outputs in Figure \ref{example-fewshot-outputs-kp20k} and Figure \ref{example-zeroshot-outputs-kp20k}. Figures \ref{example-fewshot-outputs-kp20k} presents the predictions of the low-resource models on the scientific benchmark datasets (corresponding to Table \ref{tab:scikp-main-results} and \ref{tab:scikp-main-results-pkp}). Figure \ref{example-zeroshot-outputs-kp20k} presents the predictions of zero-shot models on KP20k (corresponding to Table \ref{tab:kp20k-zeroshot-results}). We find that BART+SSR-D predicts more correct keyphrases and generally has a more diverse output.

\section{Artifact Release}
The KP20k dataset and the Fairseq library we use are MIT licensed. While commercial use is allowed for these artifacts, we only use them for research. For reproducibility, we release the three small KP20k subsets that we use as $D_{kp}$ and the code to reproduce our experiments. We refer to their original hosts for the entire training, validation, and testing datasets. In addition, we release the raw predictions of our BART+SSR-D model trained on 20k data from KP20k. Our code, data, and model outputs are released at \url{https://github.com/xiaowu0162/low-resource-kpgen}.

\begin{figure*}[h!]
\small
\centering
\begin{tabular}{p{0.98\linewidth}}
    \hline  
    \textbf{Title:} \textcolor{blue}{short signatures} from the weil \textcolor{blue}{pairing} . \\
    \textbf{Abstract:} we introduce a \textcolor{blue}{short signature} scheme based on the computational diffiehellman assumption on certain elliptic and hyperelliptic curves . for standard security parameters , the signature length is about half that of a dsa signature with a similar level of security . our \textcolor{blue}{short signature} scheme is designed for systems where signatures are typed in by a human or are sent over a low bandwidth channel . we survey a number of properties of our signature scheme such as signature aggregation and batch verification . \\
    \textbf{Ground Truth:} \textcolor{blue}{short signatures}, \textcolor{blue}{pairings}, \textcolor{blue}{bilinear maps}, \textcolor{blue}{digital signatures}, \textcolor{blue}{elliptic curves}\\
    \textbf{ExHiRD-h:} \textcolor{blue}{short signature}, weil pairing, signature aggregation, elliptic security \\
    \textbf{One2Set:} weil pairing, security, hyperelliptic signature, weil signature \\
    \textbf{BART+SSR-D:} \textcolor{blue}{short signatures}, \textcolor{blue}{pairing}, hyperelliptic curve, \textcolor{blue}{elliptic curve}, \textcolor{blue}{digital signatures} \\
    \hline  
    \textbf{Title:} computing smallest singular triplets with \textcolor{blue}{implicitly restarted} \textcolor{blue}{lanczos bidiagonalization} . \\
    \textbf{Abstract:} a matrix free algorithm , <unk> , for the efficient computation of the smallest singular triplets of large and possibly sparse matrices is described . key characteristics of the approach are its use of \textcolor{blue}{lanczos bidiagonalization} , \textcolor{blue}{implicit restarting} , and \textcolor{blue}{harmonic ritz values} . the algorithm also uses a \textcolor{blue}{deflation} strategy that can be applied directly on \textcolor{blue}{lanczos bidiagonalization} . a refinement postprocessing phase is applied to the converged singular vectors . the computational costs of the above techniques are kept small as they make direct use of the bidiagonal form obtained in the course of the \textcolor{blue}{lanczos factorization} . several numerical experiments with the method are presented that illustrate its effectiveness and indicate that it performs well compared to existing codes . \\
    \textbf{Ground Truth:} \textcolor{blue}{lanczos bidiagonalization}, \textcolor{blue}{implicit restarting}, \textcolor{blue}{harmonic ritz values}, \textcolor{blue}{deflation}, \textcolor{blue}{pseudospectrum}, \textcolor{blue}{refined singular vectors} \\
    \textbf{ExHiRD-h:} singular triplets, implicitly restarted lanczos bidiagonalization, refinement postprocessing, bidiagonalization bidiagonalization \\
    \textbf{One2Set:} singular computing, matrix triplets \\
    \textbf{BART+SSR-D:} \textcolor{blue}{lanczos bidiagonalization}, lanczos factorization, \textcolor{blue}{deflation}, matrix free algorithms, matrix eigenvalue problems \\
    \hline  
    \textbf{Title:} self stabilizing \textcolor{blue}{clock synchronization} in the presence of byzantine faults . \\
    \textbf{Abstract:} we initiate a study of bounded \textcolor{blue}{clock synchronization} under a more severe fault model than that proposed by lamport and melliar smith [digit] . realistic aspects of the problem of synchronizing clocks in the presence of faults are considered . one aspect is that clock synchronization is an on going task , thus the assumption that some of the processors never fail is too optimistic . to cope with this reality , we suggest \textcolor{blue}{self stabilizing} protocols that stabilize in any ( long enough ) period in which less than a third of the processors are faulty . another aspect is that the clock value of each processor is bounded . a single transient fault may cause the clock to reach the upper bound . therefore , we suggest a bounded clock that wraps around when appropriate . we present two randomized self stabilizing protocols for synchronizing bounded clocks in the presence of byzantine processor failures . the first protocol assumes that processors have a common pulse , while the second protocol does not . a new type of distributed counter based on the chinese remainder theorem is used as part of the first protocol . \\
    \textbf{Ground Truth:} \textcolor{blue}{self stabilization}, \textcolor{blue}{clock synchronization}, \textcolor{blue}{byzantine failures} \\
    \textbf{ExHiRD-h:} \textcolor{blue}{self stabilizing}, \textcolor{blue}{clock synchronization}, chinese remainder theorem \\
    \textbf{One2Set:} fault synchronization, synchronization, fault tolerance, bounded presence, clock reality \\
    \textbf{BART+SSR-D:} \textcolor{blue}{self stabilization}, \textcolor{blue}{clock synchronization}, byzantine faults, distributed algorithms \\
    \hline  
    \textbf{Title:} \textcolor{blue}{distributed representations} , \textcolor{blue}{simple recurrent networks} , and \textcolor{blue}{grammatical structure} . \\
    \textbf{Abstract:} in this paper three problems for a connectionist account of language are considered [digit] . what is the nature of linguistic representations [digit] . how can complex structural relationships such as constituent structure be represented [digit] . how can the apparently open ended nature of language be accommodated by a fixed resource system using a prediction task , a \textcolor{blue}{simple recurrent network} ( srn ) is trained on <unk> sentences which contain multiply embedded relative clauses . principal component analysis of the hidden unit activation patterns reveals that the network solves the task by developing complex \textcolor{blue}{distributed representations} which encode the relevant grammatical relations and hierarchical constituent structure . differences between the srn state representations and the more traditional pushdown store are discussed in the final section . \\
    \textbf{Ground Truth:} \textcolor{blue}{distributed representations}, \textcolor{blue}{simple recurrent networks}, \textcolor{blue}{grammatical structure} \\
    \textbf{ExHiRD-h:} \textcolor{blue}{distributed representations}, recurrent networks, grammatical relations, hierarchical constituent structure, hierarchical representations \\
    \textbf{One2Set:} recurrent networks, linguistic structure \\
    \textbf{BART+SSR-D:} \textcolor{blue}{distributed representations}, \textcolor{blue}{simple recurrent networks}, \textcolor{blue}{grammatical structure}, language, recurrent networks, hidden units, connectionist systems \\
    \hline  
    \textbf{Title:} \textcolor{blue}{random walks} in weyl chambers and the decomposition of \textcolor{blue}{tensor powers} . \\
    \textbf{Abstract:} we consider a class of \textcolor{blue}{random walks} on a lattice , introduced by gessel and zeilberger , for which the reflection principle can be used to count the number of k step walks between two points which stay within a chamber of a \textcolor{blue}{weyl group} . we prove three independent results about such reflectable walks first , a classification of all such walks semi second , many determinant formulas for walk numbers and their generating functions semi third , an equality between the walk numbers and the multiplicities of irreducibles in the kth \textcolor{blue}{tensor power} of certain lie group representations associated to the walk types . our results apply to the defining representations of the classical groups , as well as some spin representations of the orthogonal groups . \\
    \textbf{Ground Truth:} \textcolor{blue}{random walk}, \textcolor{blue}{tensor power}, \textcolor{blue}{weyl group}, \textcolor{blue}{hyperbolic bessel function}, \textcolor{blue}{representation of lie group} \\
    \textbf{ExHiRD-h:} \textcolor{blue}{random walks}, weyl chambers, \textcolor{blue}{tensor powers}, weyl \\
    \textbf{One2Set:} lattice chambers \\
    \textbf{BART+SSR-D:} \textcolor{blue}{random walks}, reflection principle, \textcolor{blue}{tensor powers}, lie groups, \textcolor{blue}{weyl groups}, determinant formulas, orthogonal groups, group representations, tensor product, group integrals \\
    \hline
\end{tabular}
\caption{Example outputs from low-resource models on the scentific benchmarks. The models are trained on a training set of size 20,000. Correct keyphrases are colored in \textcolor{blue}{blue}. We observe that BART+SSR-D has significantly more correct outputs and is able to predict more diverse keyphrases.}
\label{example-fewshot-outputs-kp20k}
\end{figure*}

\begin{figure*}[h!]
\small
\centering
\begin{tabular}{p{0.98\linewidth}}
    \hline
    \textbf{Title:} \textcolor{blue}{shot change detection} using scene based constraint . \\
    \textbf{Abstract:} a key step for managing a large video database is to partition the video sequences into shots . past approaches to this problem tend to confuse gradual shot changes with changes caused by smooth camera motions . this is in part due to the fact that camera motion has not been dealt with in a more fundamental way . we propose an approach that is based on a physical constraint used in \textcolor{blue}{optical flow} analysis , namely , the total brightness of a scene point across two frames should remain constant if the change across two frames is a result of smooth camera motion . since the brightness constraint would be violated across a shot change , the detection can be based on detecting the violation of this constraint . it is robust because it uses only the qualitative aspect of the brightness constraint detecting a scene change rather than estimating the scene itself . moreover , by tapping on the significant know how in using this constraint , the algorithm 's robustness is further enhanced . experimental results are presented to demonstrate the performance of various algorithms . it was shown that our algorithm is less likely to interpret gradual camera motions as shot changes , resulting in a significantly better precision performance than most other algorithms . \\
    \textbf{Ground Truth:} \textcolor{blue}{shot change detection}, \textcolor{blue}{optical flow}, \textcolor{blue}{video segmentation} \\
    \textbf{BART+TI:} cameras, computers and the internet \\
    \textbf{BART+TG:} video, computers and the internet \\
    \textbf{BART+SSR-D:} camera, \textcolor{blue}{optical flow}, video shot change detection \\
    \hline
    \textbf{Title:} a generic \textcolor{blue}{sampling framework} for improving \textcolor{blue}{anomaly detection} in the \textcolor{blue}{next generation network} . \\
    \textbf{Abstract:} the heterogeneous nature of network traffic in \textcolor{blue}{next generation networks} ( ngns ) may impose \textcolor{blue}{scalability} issue to traffic monitoring applications . while this issue can be well addressed by existing sampling approaches , owing to their inherent ' lossy ' characteristic and data reduction principle , traditional sampling techniques suffer from incomplete traffic statistics , which can lead to inaccurate inferences of the network traffic . by focusing on two distinct traffic monitoring applications , namely , \textcolor{blue}{anomaly detection} and \textcolor{blue}{traffic measurement} , we highlight the possibility of addressing the \textcolor{blue}{accuracy} of both applications without having to sacrifice one for the sake of the other . in light of this , we propose a generic \textcolor{blue}{sampling framework} , which is capable of providing creditable network traffic statistics for accurate \textcolor{blue}{anomaly detection} in the non , while at the same time preserves the principal purpose of sampling ( i.e. , to sample dominant traffic flows for accurate \textcolor{blue}{traffic measurement} ) , and thus addressing the \textcolor{blue}{accuracy} of both applications concurrently . with the emphasize on the \textcolor{blue}{accuracy} of \textcolor{blue}{anomaly detection} and the \textcolor{blue}{scalability} of monitoring devices , the performance evaluation over real network traces demonstrates the superiority of the proposed framework over traditional sampling techniques . copyright ( c ) [digit] john wiley sons , ltd . \\
    \textbf{Ground Truth:} \textcolor{blue}{sampling framework}, \textcolor{blue}{anomaly detection}, \textcolor{blue}{next generation network}, \textcolor{blue}{scalability}, \textcolor{blue}{traffic measurement}, \textcolor{blue}{accuracy} \\
    \textbf{BART+TI:} ngns, computers and the internet, tech industry \\
    \textbf{BART+TG:} computers and the internet, wireless communications \\
    \textbf{BART+SSR-D:} ngn, \textcolor{blue}{anomaly detection}, \textcolor{blue}{traffic measurement}, wireless, nsa \\    
    \hline
    \textbf{Title:} recent developments in \textcolor{blue}{high level synthesis} . \\
    \textbf{Abstract:} we survey recent developments in \textcolor{blue}{high level synthesis} technology for \textcolor{blue}{vlsi design} . the need for higher level \textcolor{blue}{design automation} tools are discussed first . we then describe some basic techniques for various subtasks of \textcolor{blue}{high level synthesis} . techniques that have been proposed in the past few years ( since [digit] ) for various subtasks of \textcolor{blue}{high level synthesis} are surveyed . we also survey some new synthesis objectives including testability , power efficiency , and reliability . \\
    \textbf{Ground Truth:} \textcolor{blue}{high level synthesis}, \textcolor{blue}{vlsi design}, \textcolor{blue}{design automation}, \textcolor{blue}{design methodology} \\
    \textbf{BART+TI:} design, computers and the internet \\
    \textbf{BART+TG:} design, computers and the internet \\
    \textbf{BART+SSR-D:} \textcolor{blue}{high level synthesis}, vlsi \\
    \hline
    \textbf{Title:} \textcolor{blue}{asynchronous parallel finite automaton} a new mechanism for deep packet inspection in \textcolor{blue}{cloud computing} . \\
    \textbf{Abstract:} security is quite an important issue in \textcolor{blue}{cloud computing} . the general security mechanisms applied in the cloud are always passive defense methods such as encryption . besides these , it 's necessary to utilize real time active monitoring , detection and defense technologies . according to the published researches , \textcolor{blue}{deep packets inspection} ( dpi ) is the most effective technology to realize active inspection and defense . however , most of the works on dpi focus on its performance in general application scenarios and make improvement for space reduction , which could not meet the demands of high speed and stability in the cloud . therefore it is meaningful to improve the common mechanisms of dpi , making it more suitable for \textcolor{blue}{cloud computing} . in this paper , an \textcolor{blue}{asynchronous parallel finite automaton} ( fa ) is proposed . the applying of asynchronous parallelization and heuristic forecast mechanism decreases the time consumed in matching significantly , while still reduces the memory required . moreover , it is immune to overlapping problem , also enhancing the stability . the final evaluation results show that asynchronous parallel fa has higher stability , better performance on both time and memory , and is more suitable for \textcolor{blue}{cloud computing} . \\
    \textbf{Ground Truth:} \textcolor{blue}{asynchronous parallel finite automaton}, \textcolor{blue}{deep packet inspection}, \textcolor{blue}{cloud computing}, \textcolor{blue}{lock free fifo} \\
    \textbf{BART+TI:} \textcolor{blue}{cloud computing}, computer security \\
    \textbf{BART+TG:} \textcolor{blue}{cloud computing}, dpi \\
    \textbf{BART+SSR-D:} \textcolor{blue}{cloud computing}, parallel finite automaton, deep packets inspection ( dpi ), computer security \\
    \hline 
\end{tabular}
\caption{Example zero-shot cross-domain transfer outputs on the scentific benchmarks. We train the models with KP20k as $D_{aux}$ and KPTimes as $D_{kp}$. Correct keyphrases are colored in \textcolor{blue}{blue}. We observe that the KPTimes model fine-tuned on BART+SSR-D is able to predict sinificantly more diverse and relevant keyphrases. It also has some correct predictions while BART+TI or BART+TG barely have any. }
\label{example-zeroshot-outputs-kp20k}
\end{figure*}

\end{document}